\documentclass[lettersize,journal]{IEEEtran}
\usepackage{amsmath,amsfonts}
\usepackage{algorithmic}
\usepackage{algorithm}
\usepackage{array}
\usepackage[caption=false,font=normalsize,labelfont=sf,textfont=sf]{subfig}
\usepackage{textcomp}
\usepackage{stfloats}
\usepackage{url}
\usepackage{verbatim}
\usepackage{graphicx}
\usepackage{cite}
\usepackage{multirow,color}
\usepackage{tabularx}
\usepackage{hhline}
\usepackage{booktabs}
\usepackage{times} 

\begin{document}

\title{Why Look at It at All?: Vision-Free Multifingered Blind Grasping Using Uniaxial Fingertip Force Sensing}

\author{Edgar~Lee,  Junho~Choi,  Taemin~Kim,  Changjoo Nam*,  and~Seokhwan~Jeong*
\thanks{This work was supported by the National Research Foundation of Korea (NRF) grant funded by the Korea government (MSIT) (No. RS-2024-00461583) and in part by the
Nano \& Material Technology Development Program through the NRF funded by the Ministry of Science and ICT under Grant (RS-2025-25442536)
} 
\thanks{E. Lee, J. Choi, T. Kim and S. Jeong are with the Department of Mechanical Engineering, Sogang University, Seoul, South Korea.}
\thanks{C. Nam is with the Department of Electronic Engineering, Sogang University, Seoul, South Korea.}
\thanks{C. Nam and S. Jeong are the corresponding authors (email: cjnam@sogang.ac.kr, seokhwan@sogang.ac.kr)}
}

\markboth{Journal of \LaTeX\ Class Files,~Vol.~xx, No.~x, December~2025}%
{Shell \MakeLowercase{\textit{et al.}}: A Sample Article Using IEEEtran.cls for IEEE Journals}


\maketitle

\begin{abstract}
Grasping under limited sensing remains a fundamental challenge for real-world robotic manipulation, as vision and high-resolution tactile sensors often introduce cost, fragility, and integration complexity. This work demonstrates that reliable multifingered grasping can be achieved under extremely minimal sensing by relying solely on uniaxial fingertip force feedback and joint proprioception, without vision or multi-axis/tactile sensing. To enable such blind grasping, we employ an efficient teacher-student training pipeline in which a reinforcement-learned teacher exploits privileged simulation-only observations to generate demonstrations for distilling a transformer-based student policy operating under partial observation. The student policy is trained to act using only sensing modalities available at real-world deployment. We validate the proposed approach on real hardware across 18 objects, including both in-distribution and out-of-distribution cases, achieving a 98.3~$\%$ overall grasp success rate. These results demonstrate strong robustness and generalization beyond the simulation training distribution, while significantly reducing sensing requirements for real-world grasping systems.

\end{abstract}


\section{Introduction}

\IEEEPARstart{R}{ecent} advances in learning-based techniques have significantly accelerated research on multifingered robotic manipulation by leveraging diverse sensor modalities, such as vision, tactile, force, visual-tactile, and proprioceptive sensing~\cite{andrychowicz2020learning,chen2023visual,singh2024dextrah,qi2023general,suresh2024neuralfeels,liang2021multifingered,pitz2023dextrous}. These methods enable robots to perform dexterous and adaptive manipulation, and they are increasingly used in real robotic platforms.

\par While advanced multimodal systems enable highly dexterous manipulation, their cost can limit scalability and real-world adoption. To reduce sensing cost, recent work explores simplified modalities, including single RGB-D cameras as low-cost alternatives to multi-camera vision~\cite{chen2023visual}, low-cost visual-tactile sensors (e.g., DIGIT) and plug-and-play tactile skins (e.g., AnySkin) that simplify integration and offer streamlined fabrication~\cite{lambeta2020digit,bhirangi2025anyskin}, and FSR-based fingertip sensors as substitutes for expensive multi-axis force/torque sensing in manipulation pipelines~\cite{yin2023rotating,lee2024dextouch,yuan2024robot}.

\begin{figure}
\centering
\includegraphics[width=8.75cm]{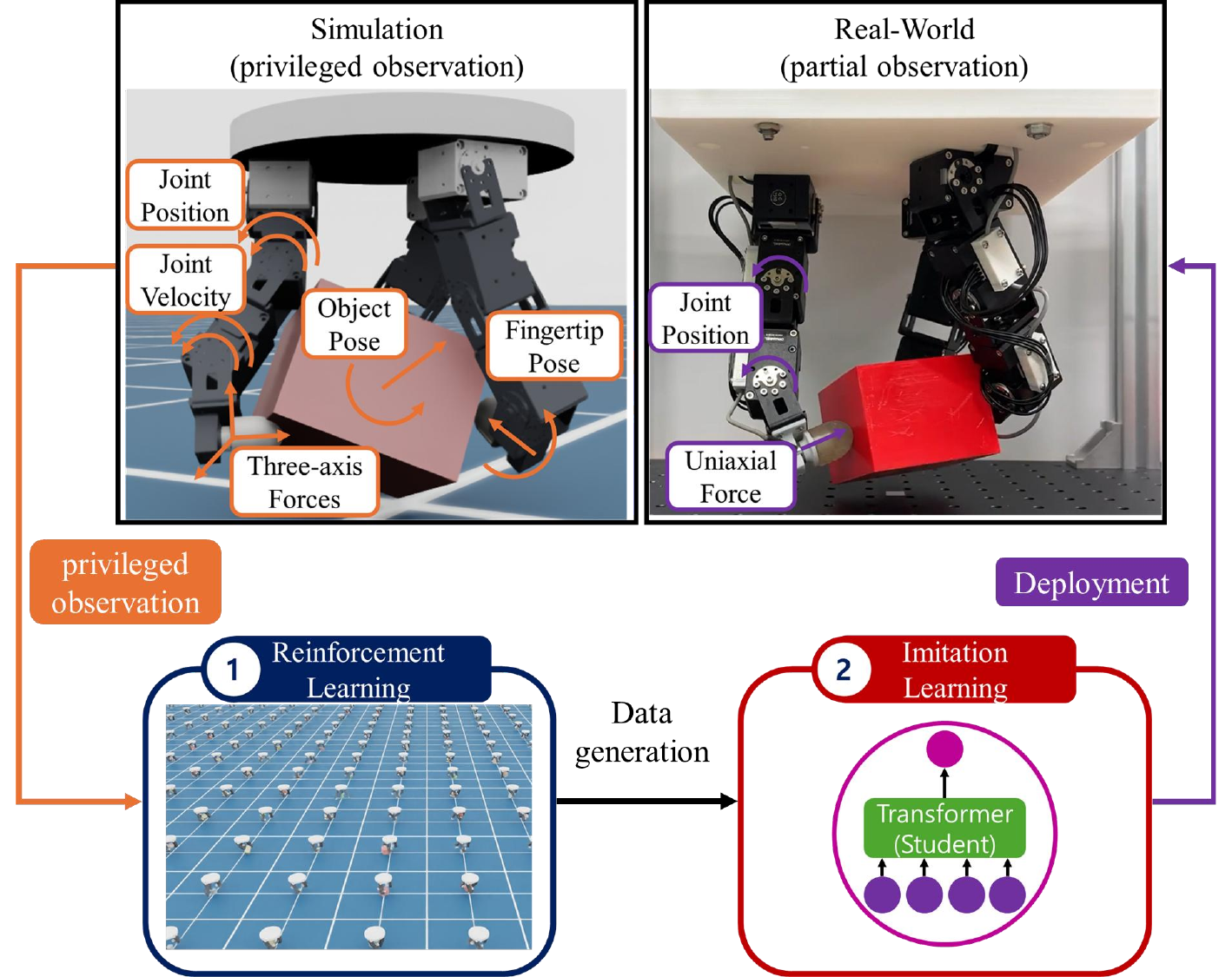}
\caption{Overview of the proposed teacher-student training pipeline for blind grasping.}
\label{abstract}
    \vspace{-0.65cm}
\end{figure}

\par Despite these efforts, each modality still presents unresolved challenges. Vision-based sensing is inexpensive, yet occlusions and lighting variation degrade object-state estimation, often motivating domain randomization or visual data augmentation during training~\cite{xia2024targo,andrychowicz2020learning,singh2024dextrah}. Tactile sensing provides rich local contact information, but visual-tactile sensors with optical paths and elastomer skins are difficult to miniaturize and remain sensitive to illumination/reflectance, while elastomer geometry can also constrain fingertip contact modalities~\cite{lambeta2020digit,suresh2024neuralfeels}. Soft-type tactile sensors better conform to fingertip shapes but are still largely at a laboratory stage~\cite{lin2025learning,im2025simultaneous}. Force sensing is easy to mount, but multi-axis sensors are expensive and hard to miniaturize, and low-cost FSRs suffer from nonlinearity, hysteresis, and poor repeatability, limiting them to coarse contact detection rather than precise force estimation. To address these limitations, recent research has explored multimodal visual-tactile sensing~\cite{qi2023general,suresh2024neuralfeels,yuan2024robot}. These sensors can improve robustness but often inherit bulkiness, fragility, and internal-lighting sensitivity, hindering widespread industrial adoption. Consequently, an intermediate sensing paradigm is needed for \textit{vision-free} (i.e., blind) and contact-based grasping tasks in industrial settings that is low-cost, robust, simple, geometry-flexible, and unaffected by visual conditions.

\par In this paper, we introduce a blind grasping framework driven by a uniaxial force sensor, along with a teacher-student training pipeline for efficient learning (see Fig.~\ref{abstract}). The proposed sensor enables precise and continuous force estimation with a robust, low-cost, and compact hardware design. It supports diverse fingertip attachments for stable contact and is fully independent of visual conditions. The proposed formulation also simplifies simulation by removing the need for photorealistic rendering and additional tactile simulation modules. By operating solely on uniaxial force sensing and proprioceptive signals (e.g., joint states), our approach eliminates auxiliary perception tasks such as segmentation, depth estimation, and 6D pose estimation. The main contributions of this paper are summarized as follows:

\begin{itemize}
  \item We demonstrate that reliable multifingered grasping can be achieved using extremely minimal sensing, relying solely on uniaxial fingertip force feedback and joint proprioception, without vision or multi-axis tactile sensing. This significantly lowers sensing and integration requirements for real-world deployment.
  
  \item We develop a teacher--student training framework tailored for blind grasping, where a reinforcement-learned teacher uses privileged simulation-only observations with tailored rewards to encourage reliance on deployable partial observations, and distills a transformer student from curated demonstrations of successful grasps with rich uniaxial force interactions.
  
  \item We validate the proposed approach on real hardware across 18 objects (6 in-distribution and 12 out-of-distribution), achieving a 98.3~$\%$ overall grasp success rate, thereby demonstrating strong robustness and generalization beyond the simulation training distribution.
\end{itemize}

\par The real-world hardware comprises a custom three-fingered gripper mounted on a stationary frame (see Fig.~\ref{abstract}). The gripper is derived from the open-source D’Claw manipulator~\cite{Kumar_ROBEL}, with three 3-DoF fingers (9-DoF in total) driven by servo motors (Dynamixel XM-430, Robotis). Each fingertip is equipped with a uniaxial force sensor. The supplementary video provides real-world demonstrations of the proposed system.

\section{RELATED WORK} \label{sec:Related}
\par \textbf{Multifingered Manipulation:} Multifingered manipulation has evolved with the development of diverse sensor modalities. Classical methods relied on analytical modeling and precise sensing to achieve dexterous control. In~\cite{park2021softness}, fingertip forces were estimated using multiple custom tactile sensors mounted on fingertips to grasp objects according to rule-based softness-adaptive pinch-grasp algorithm, and in~\cite{shaw2018tactile}, tactile feedback was leveraged in a model-based, vision-free framework for manipulating novel objects. While effective, these approaches often depend on expensive tactile hardware and highly accurate models, limiting their scalability to real-world deployment. Hybrid approaches that combine analytical methods with learning or motion planning have also been studied. For example,~\cite{grace2024direct} uses particle filtering to identify an inverse Jacobian for in-hand manipulation, and ~\cite{pang2023global} employs a quasi-dynamic model with smoothed contact dynamics for motion-planning-based manipulation. However, such model-dependent frameworks are typically confined to pre-defined contact tasks and require accurate dynamics that are difficult to obtain in practice.

\par Recent studies have focused on learning-based visuomotor manipulation, leveraging visual sensing for dexterous reorientation and grasping via multiple tracking cameras and RGB inputs~\cite{andrychowicz2020learning}, single RGB-D for cost-effective in-hand reorientation~\cite{chen2023visual}, or dual RGB-D inputs for joint arm-hand control across diverse objects~\cite{singh2024dextrah}. However, vision-only pipelines are vulnerable to occlusions, inconsistent visibility, and lighting variations, making object-state estimation unreliable in real-world settings. To address these limitations, prior work explores multimodal fusion for robust in-hand manipulation. Specifically,~\cite{qi2023general} integrates proprioceptive positions with visual and tactile feedback for multi-axis object rotation,~\cite{suresh2024neuralfeels} fuses visual and tactile sensing to estimate object pose and shape during in-hand manipulation, and~\cite{yuan2024robot} combines visual cues with FSR-based feedback to improve manipulation accuracy.

\par Beyond multimodal fusion, recent efforts explore vision-free (blind) manipulation, where control relies solely on proprioceptive or force feedback. Prior work leverages joint angles, tactile contacts, and discretized torque levels to learn multifingered grasping~\cite{liang2021multifingered}, torque feedback for object reorientation~\cite{pitz2023dextrous}, proprioceptive/action history for fingertip-based object rotation~\cite{qi2023hand} and for manipulating elongated objects along the $z$-axis~\cite{wang2024lessons}, and a tactile-based sim-to-real approach that seeks and manipulates randomly placed targets in the dark (e.g., grasping, door opening, and valve rotation) using binary contact feedback and a coarse target-range prior~\cite{lee2024dextouch}.


\par \textbf{Learning Approaches:} From a learning perspective, robotic policy training is primarily studied under two paradigms: reinforcement learning (RL) and imitation learning (IL). RL optimizes a policy through simulated interactions with designed rewards~\cite{tan2018sim, rudin2022learning}, but often requires extensive iterations and rich visual observations~\cite{chen2023visual,akkaya2019solving}, making it computationally demanding. In contrast, IL optimizes a policy from expert demonstrations~\cite{mandlekar2021matters}, offering data efficiency and intuitive supervision, yet it typically demands manual data collection~\cite{zhao2023learning, lin2025learning} and can be sensitive to unreliable human demonstrations that degrade generalization~\cite{chen2025curating}. 

\par More recently, the teacher-student framework has been proposed to bridge the gap between privileged and partial observations. In~\cite{chen2020learning}, a teacher trained with full sensory inputs supervises a student constrained to partial observations, improving sample efficiency over training from scratch. Several works adopt a two-stage pipeline that trains a teacher with RL and distills it into a student, e.g., via behavioral cloning~\cite{bauza2024demostart} or DAgger~\cite{ross2011reduction}-based learning as in~\cite{chen2023visual}, further refined with an online DAgger variant~\cite{singh2024dextrah}. Temporal context has also been explored: history-based proprioceptive sequences improve action prediction~\cite{wang2024lessons}, and transformer architecture captures long-horizon dependencies among proprioceptive states and actions~\cite{chen2024vegetable}.

\section{LEARNING METHOD}  \label{sec:Algorithm}

This section presents a teacher-student framework for training a blind grasping policy under partial observations. A teacher is trained with privileged observations via RL, then distilled into a student via IL, yielding a final policy that uses only proprioceptive position and uniaxial force inputs for blind dexterous grasping. All training is conducted in IsaacLab~\cite{mittal2023orbit}, which supports large-scale parallel simulation.

\begin{figure}
\centering
\includegraphics[width=8.75cm]{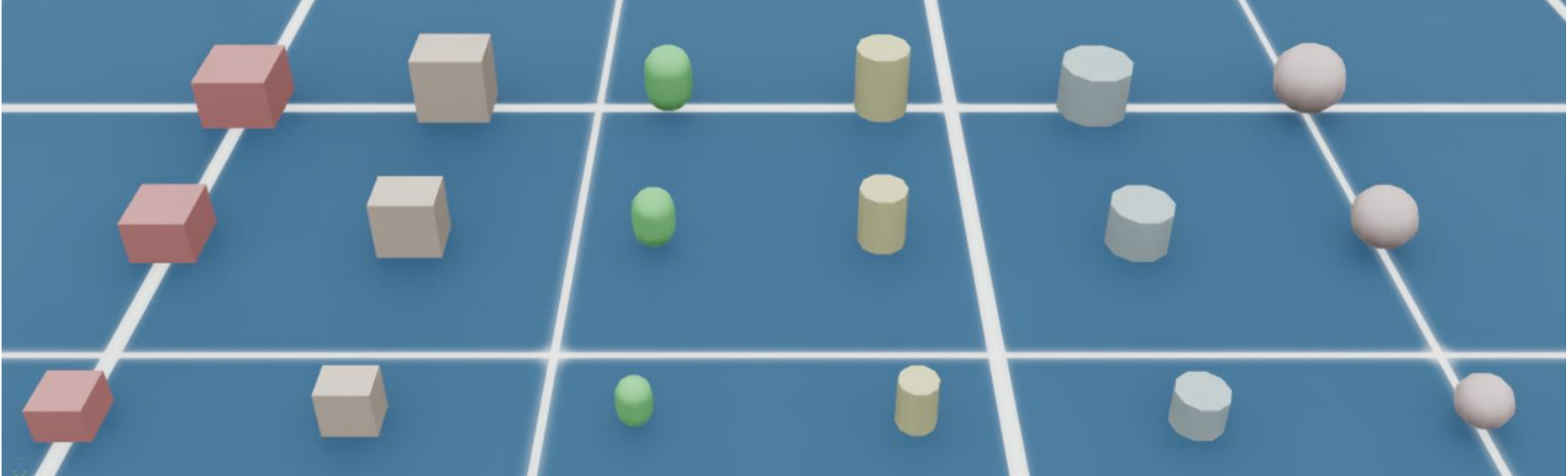}
\caption{Set of the 18 geometric objects used in simulation for training the teacher policy. The object set comprises six geometries (two cuboids A/B, one capsule, two cylinders A/B, and one sphere), each modeled in three sizes.}

\label{sim_object}
\vspace{-0.45cm}
\end{figure}

\begin{table}[!t]
\caption{Dimensions of the 18 geometric objects used in simulation\label{tab:object_size}}
\centering
\resizebox{\columnwidth}{!}{%
\begin{tabular}{lccccc}
\toprule
\textbf{Geometry} & \textbf{Parameter} & \textbf{Large} & \textbf{Medium} & \textbf{Small} & \textbf{Unit} \\
\midrule
\multirow{3}{*}{Cuboid A} & X & 10.0 & 8.0 & 6.0 & cm \\
                          & Y & 10.0 & 8.0 & 6.0 & cm \\
                          & Z & 7.5 & 6.0 & 4.5 & cm \\
\midrule
\multirow{3}{*}{Cuboid B} & X & 10.0 & 8.0 & 6.0 & cm \\
                          & Y & 7.5 & 6.0 & 4.5 & cm \\
                          & Z & 10.0 & 8.0 & 6.0 & cm \\
\midrule
\multirow{2}{*}{Capsule} & Radius & 3.33 & 2.67 & 2.00 & cm \\
                         & Height & 3.33 & 2.67 & 2.00 & cm \\
\midrule
\multirow{2}{*}{Cylinder A} & Radius & 3.75 & 3.0 & 2.25 & cm \\
                            & Height & 10.0 & 8.0 & 6.0 & cm \\
\midrule
\multirow{2}{*}{Cylinder B} & Radius & 5.0 & 4.0 & 3.0 & cm \\
                            & Height & 7.5 & 6.0 & 4.5 & cm \\
\midrule
\multirow{1}{*}{Sphere} & Radius & 5.0 & 4.0 & 3.0 & cm \\
\bottomrule

\end{tabular}%
}
\vspace{-0.5cm}
\end{table}

\subsection{Teacher Policy Training} \label{sec:Teacher}
\par The teacher policy $\pi_{t}$ is trained in simulation with privileged observations to perform blind grasping from an initial non-contact state. The task is to make contact, establish a grasp, and smoothly lift the object. The training set contains the 18 geometric objects (one sphere, two cuboids, one capsule, and two cylinders), each in three sizes that fit the gripper workspace (see Table~\ref{tab:object_size}, Fig.~\ref{sim_object}).

\par For sample-efficient and stable RL, the teacher receives a privileged observation $o_t^{\text{priv}}\in\mathbb{R}^{95}$ that better satisfies the Markov assumption~\cite{sutton2018reinforcement}. This vector includes joint positions/velocities, fingertip pose, object pose, object linear/angular velocities, three-axis fingertip contact forces, the planar distance between the object and gripper center, uniaxial fingertip forces, 6D fingertip wrench (force and torque) and the previous action. Gaussian noise is added to improve robustness and sim-to-real transfer ($\sigma=0.005~\text{rad}$ for joint angles and $\sigma=0.5~\text{N}$ for fingertip forces). Each uniaxial force is computed by projecting the 3-axis fingertip contact forces onto the fingertip local \( z\text{-axis} \) and thresholding small magnitudes to suppress noise.

\begin{figure}
\centering
\includegraphics[width=8.75cm]{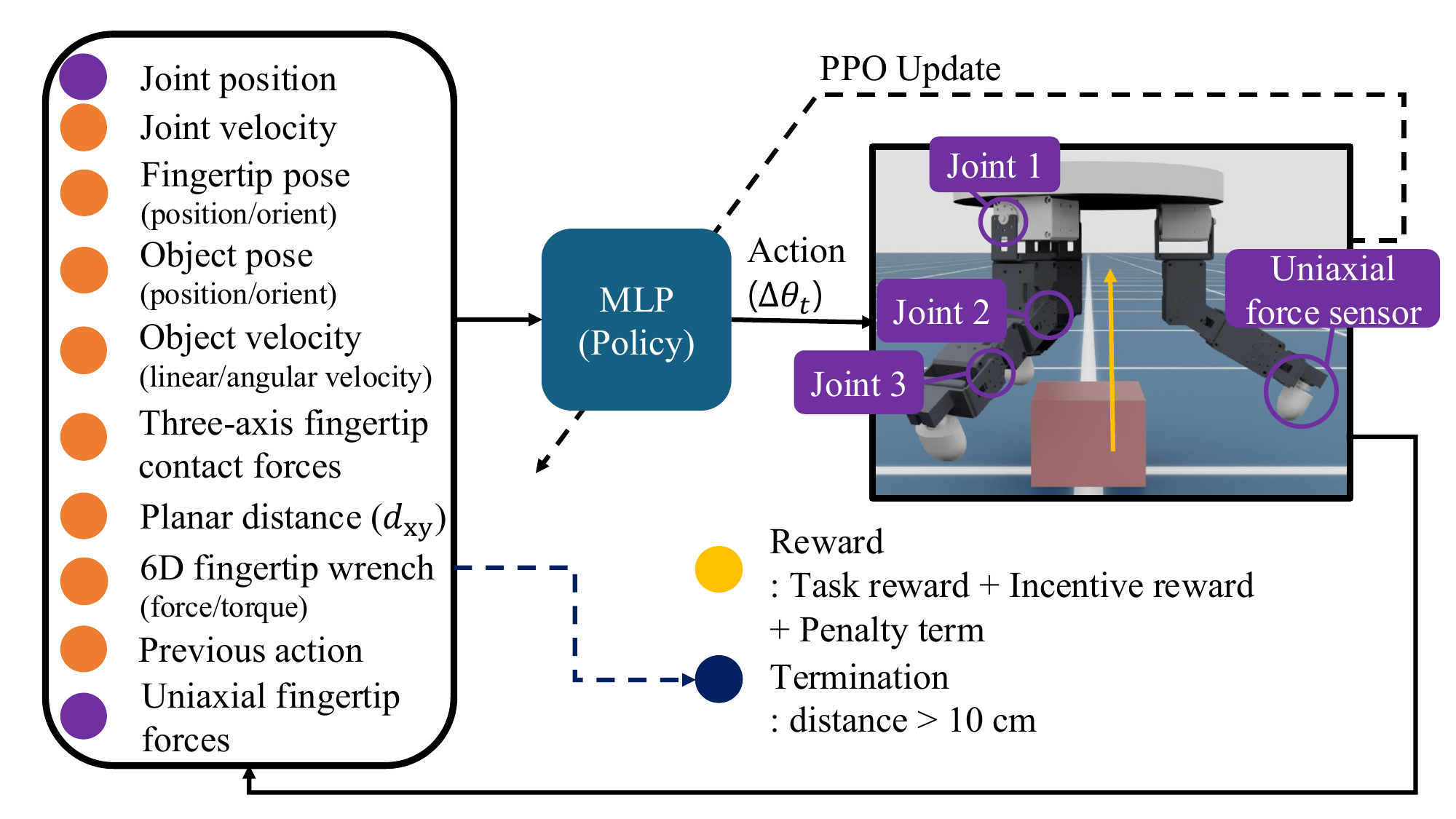}

\caption{Pipeline of the teacher policy trained in simulation with privileged observations. An MLP takes privileged observations and outputs relative joint position changes ($\Delta\theta_t$). Each finger comprises three actuated joints and a uniaxial fingertip force sensor. The reward function includes task, incentive, and penalty terms. An episode terminates when the object distance exceeds $10~\text{cm}$ in the $xy$-plane.}

\label{teacher_policy}
    \vspace{-0.65cm}
\end{figure}

\par The policy outputs relative joint position changes $a_t=\Delta\theta_t\in\mathbb{R}^9$ at 20~Hz, clipped to [$-0.3$~rad, $0.3$~rad] only to suppress unrealistic action magnitudes while allowing rapid joint position changes. Episodes start from an initial joint configuration ($0$, $\pi/3$, $0$ for joint~$1$-$3$ of each finger; see Fig.~\ref{teacher_policy}), with per-reset joint offsets sampled from $\mathcal{U}(-\pi/9,\pi/9)$ and object $xy$ position randomized within $\pm 3$~cm from the gripper center. An episode terminates at 10~s or when the object moves more than 10~cm away from the gripper center in the $xy$-plane.

\par The reward function is designed to promote contact initiation, stable grasping from an initial non-contact state, and smooth lifting. It consists of a task reward $r_{\text{t}}$, an incentive reward $r_i$ that encourages simultaneous three-fingertip contact using uniaxial force feedback, and penalty terms ($r_{\text{l}}$, $r_{\text{a}}$, $r_{\text{ar}}$) for joint-limit violations, large actions, and rapid action-rate changes. The reward function is expressed as follows:

\vspace{-0.4cm}
\begin{equation}
\begin{split}
r = w_{1} r_{\text{t}} + w_{2} r_{\text{i}} 
+ w_{31} r_{\text{l}} + w_{32} r_{\text{a}} + w_{33} r_{\text{ar}}
\end{split}
\label{eq:reward_function}
\end{equation}
\vspace{-0.4cm}

\noindent where $w_{1}=1.0$, $w_{2}=0.2$, $w_{31}=-500$, $w_{32}=-0.04$, $w_{33}=-0.01$. Specifically, each reward term is defined as follows.

\par  The task reward $r_{\text{t}}$ promotes contact initiation and steady lifting by increasing as the object height $h_{\text{o}}$ approaches the target height $h_{\text{t}}$ and as the planar object-gripper distance $d_{xy}$ decreases, where $d_{xy}$ is defined on the $xy$-plane between $(x_{\text{obj}},y_{\text{obj}})$ and $(x_{\text{c}},y_{\text{c}})$. The planar distance is defined as

\vspace{-0.2cm}
\begin{equation}
d_{xy} = \sqrt{(x_{\text{obj}} - x_c)^2 + (y_{\text{obj}} - y_c)^2}
\label{eq:planar_distance}
\end{equation}
\vspace{-0.4cm}

\noindent The reward is activated only when at least one fingertip is in contact with the object ($n_{\text{contact}} \ge 1$), where $n_{\text{contact}}$ represents the number of fingertip force sensors detecting a contact force greater than $0.1~\text{N}$, and the object remains within the workspace limit ($d_{xy} < 10~\text{cm}$). Otherwise, the reward is set to zero. The formulation is given by:

\vspace{-0.3cm}
\begin{equation}
r_{\text{t}} =
\begin{cases}
\left(1 - \sqrt{\dfrac{h_{t} - h_{o}}{h_t}}\right)\, \left(1 - \sqrt{\dfrac{d_{xy}}{0.1}}\right), & h_{o} < h_{t} \\[10pt]
\left(1 - \sqrt{\dfrac{d_{xy}}{0.1}}\right) & h_{o} \geq h_{t} \\[10pt]
\end{cases}
\label{eq:task_reward}
\end{equation}
\vspace{-0.3cm}

\par The incentive reward $r_i$ encourages the policy to utilize uniaxial force feedback by providing a binary signal that is active only when all three fingertips are in contact ($n_{\text{contact}}=3$):

\vspace{-0.3cm}
\begin{equation}
r_{i} =
\begin{cases}
1, & n_{\text{contact}} = 3 \\
0, & \text{otherwise}
\label{eq:incentive_reward}
\end{cases}
\end{equation}
\vspace{-0.3cm}

\par To promote safe and smooth control, we add three penalty terms. The joint-limit penalty $r_{\text{l}}$ sums joint-angle violations beyond the allowable range over all nine joints:

\vspace{-0.3cm}
\begin{equation}
r_{\text{l}} =
\sum_{j=1}^{9}
\left[
\max\!\left(0,\, \theta_j^{\min} - \theta_j \right) +
\max\!\left(0,\, \theta_j - \theta_j^{\max} \right)
\right]
\label{eq:pos_limit_penalty}
\end{equation}
\vspace{-0.3cm}

\noindent The action-magnitude penalty $r_{\text{a}}$ discourages large actions:

\vspace{-0.3cm}
\begin{equation}
r_{\text{a}} =
|a_t|_2^2 =
\sum_{k=1}^{9} (a_{t,k})^2
\label{eq:action_penalty}
\end{equation}
\vspace{-0.3cm}

\noindent The action-rate penalty $r_{\text{ar}}$ discourages abrupt changes:

\vspace{-0.3cm}
\begin{equation}
r_{\text{ar}} =
 \|a_t - a_{t-1}\|_2^2 =
\sum_{k=1}^{9} \big(a_{t,k} - a_{t-1,k}\big)^2
\label{eq:action_rate_penalty}
\end{equation}
\vspace{-0.3cm}

\begin{table}[t]
\caption{Domain Randomization parameters used in simulation\label{tab:domain_randomization}}
\centering
\begin{tabularx}{\linewidth}{>{\raggedright\arraybackslash}X >{\centering\arraybackslash}X}
\hline
\textbf{Parameter} & \textbf{Range} \\
\hline
Robot: Initial Joint Offset & $+\mathcal{U}[-\pi/9, \pi/9]$ \\
Object: Initial Position ($x$, $y$) & $[-3\,\text{cm}, 3\,\text{cm}]$ \\
\hline
Object: Static Friction & $[0.7, 1.3]$ \\
Object: Dynamic Friction & $[0.7, 1.3]$, $\leq$ Static Friction \\
Object: Mass (kg) & $\times[0.5, 0.3]$ \\
Robot: Static Friction & $[0.7, 1.3]$ \\
Robot: Dynamic Friction & $[0.7, 1.3]$, $\leq$ Static Friction \\
\hline
Actuator: P (Stiffness) Gain & $\times \mathcal{U}[\log 0.3, \log 3.0]$ \\
Actuator: D (Damping) Gain & $\times \mathcal{U}[\log 0.75, \log 1.5]$ \\
\hline
Random Force: Range & $[-0.3, 0.3]$ \\
Random Force: period (sec) & $[0, 10]$ \\
\hline
Noise: Joint Angle (rad) & $+ \mathcal{N}(0, 0.005)$ \\
Noise: Force Sensor (N) & $+ \mathcal{N}(0, 0.5)$ \\
\hline
\end{tabularx}
\vspace{-0.5cm}
\end{table}

\begin{table}[t]
\caption{Hyperparameters for teacher policy training\label{tab:optimization_hyper}}
\centering
\small
\begin{tabularx}{\linewidth}{>{\raggedright\arraybackslash}X >{\centering\arraybackslash}X}
\hline
\textbf{Hyperparameters} & \textbf{Values} \\
\hline
$\#$ environments & $9000~(18~\text{objects} \times 500)$ \\
$\#$ steps per env & $24$ \\
max iterations & $4500$ \\
activation & elu \\
$\#$ learning\_epochs & $5$ \\
$\#$ minibatches & $6$ \\
learning rate & $0.001$ \\
entropy coefficient & $0.003$ \\
value loss coefficient & $1.0$ \\
clip range & $0.2$ \\
$\gamma$ & $0.99$ \\
$\lambda$ & $0.95$ \\
desired KL & $0.01$ \\
max gradient norm & $1.0$ \\
clip actions & $0.3$ \\
\hline
\end{tabularx}
\vspace{-0.5cm}
\end{table}

\par To improve robustness and generalization, we apply domain randomization over joint offsets, object positions, object friction and mass, robot friction, actuator gains, and external disturbance forces. Joint offsets and object positions are resampled at every episode reset, whereas friction-related parameters and actuator gains are randomized once at environment initialization. During each episode, a random external force $(F_x, F_y, F_z)$ is applied to the object at randomly selected times, with each component uniformly sampled from predefined ranges. The randomization ranges are summarized in Table~\ref{tab:domain_randomization}.

\par The teacher policy is trained in IsaacLab using the proximal policy optimization (PPO)~\cite{schulman2017proximal,schwarke2025rsl}. The actor and critic networks are  multilayer perceptrons (MLP) with hidden dimensions $[512, 256, 128]$, and an entropy coefficient of $0.003$ is used to encourage exploration and stabilize optimization~\cite{ahmed2019understanding}. Training runs in $9000$ parallel environments with 18 object types evenly distributed, on a desktop with NVIDIA RTX 5080 GPU, and completes in approximately $3$ hours. The detailed hyperparameters are provided in Table~\ref{tab:optimization_hyper}.

\subsection{Student Policy Training} \label{student policy}

\begin{figure}
\centering
\includegraphics[width=8.75cm]{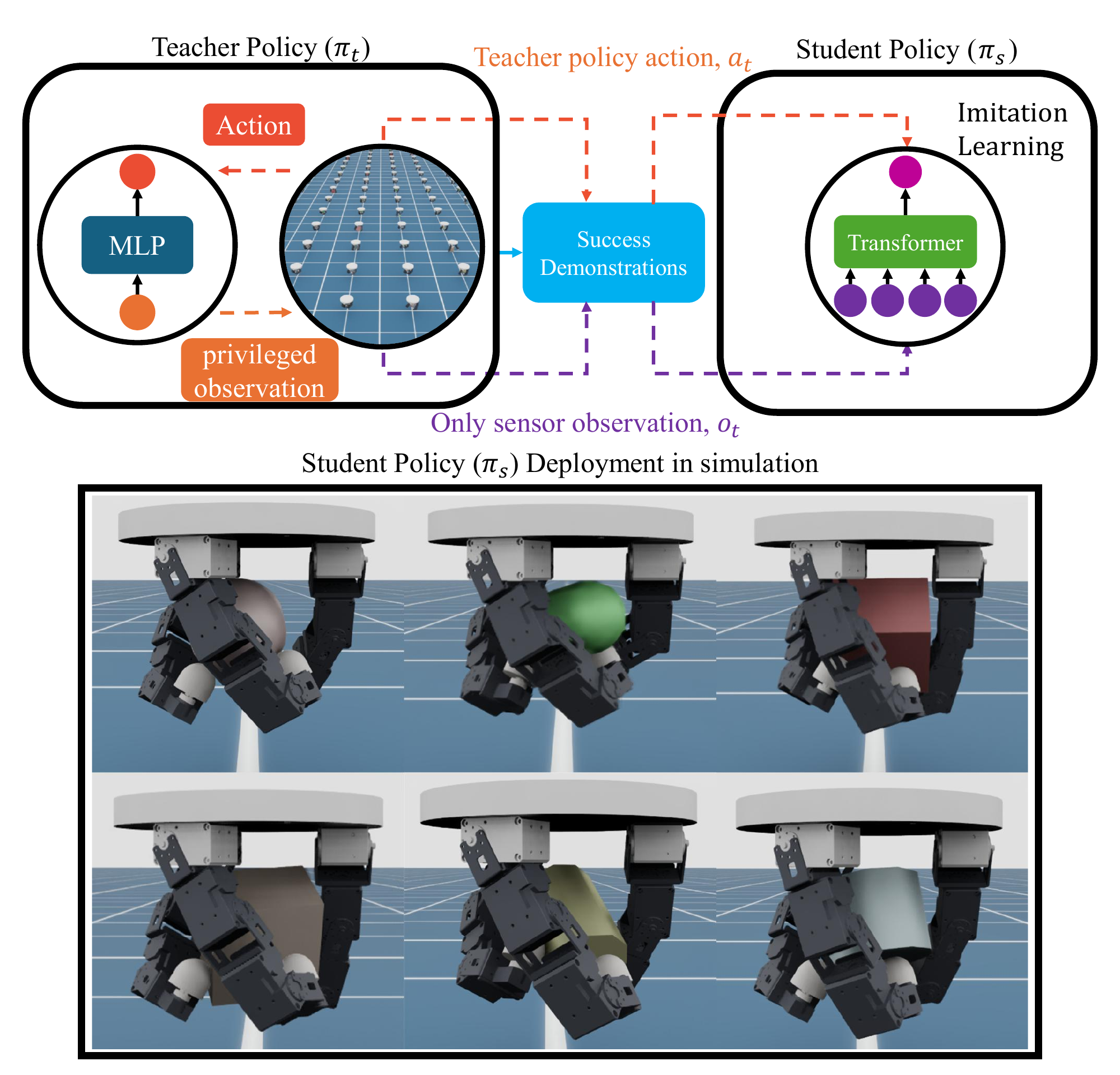}

\caption{Pipeline of the \textbf{student policy} distilled via behavioral cloning for blind grasping (top). A transformer maps proprioceptive states and uniaxial fingertip forces to relative joint position changes ($\Delta\theta_t$), trained on successful teacher demonstrations with an MSE action loss. Simulation rollouts of $\pi_s$ on in-distribution objects (bottom).}

\label{student_policy}
    \vspace{-0.45cm}
\end{figure}

\begin{table}[t]
\caption{Hyperparameters for student policy training\label{tab:transformer_hyper}}
\centering
\small
\begin{tabularx}{\linewidth}{>{\raggedright\arraybackslash}X >{\centering\arraybackslash}X}
\hline
\textbf{Hyperparameters} & \textbf{Values} \\
\hline
batch size & 256 \\
number of epochs & 2000 \\
optimizer & AdamW \\
learning rate & $1\times10^{-4}$ \\
learning rate schedule & linear decay after 100 epochs \\
L2 regularization & 0.01 \\
loss type & mean squared error (L2 loss) \\
Transformer layers & 6 \\
embedding dimension & 512 \\
number of attention heads & 8 \\
context length & 30 \\
dropout rate & 0.1 \\
activation function & GELU \\
MLP head dimension & [1024, 1024] \\
frame stacking length & 30 \\
\hline
\end{tabularx}
\vspace{-0.5cm}
\end{table}

\par We train a student policy $\pi_{s}$ via behavioral cloning (BC) to distill the teacher into a policy that operates only on partial observations consisting of proprioception and uniaxial force sensing. This stage enables blind grasping without vision/depth modalities, high-resolution/multi-axis tactile/force sensing, or explicit object-state inputs. The student observation $o_{t}\in \mathbb{R}^{12}$ includes joint positions and uniaxial fingertip forces. The overall training pipeline is illustrated in Fig.~\ref{student_policy}.

\par To collect high-quality data for distillation, we deploy the trained teacher policy $\pi_{t}$ in the same simulation setup (9000 parallel environments with identical randomization), except that external disturbance forces are disabled to avoid distorting demonstration trajectories. During rollout, the stochastic teacher is executed while teacher actions $a_t$ and student observations $o_t$ are logged at every timestep. We then curate only successful episodes based on the task reward $r_t$, keeping trajectories where $r_t>0.5$ holds continuously for at least $0.5$~s within a 5~s episode, and where the cumulative sum of uniaxial fingertip contact forces (summed across three sensors) exceeds 
$200$~N. This selection emphasizes the grasp-and-lift phase rather than the prolonged holding. Using parallel rollout, we obtain $89500$ curated demonstrations in $1$~hour. Note that these curated demonstrations are re-collected \textit{after} the teacher policy has been fully trained. Although the demonstrations are collected from the same randomization distribution used in teacher training, some newly sampled environments correspond to underrepresented scenarios that the teacher has not sufficiently learned. Consequently, these environments either fail to satisfy the success condition or require significantly longer interaction time to achieve success. To maintain data collection efficiency, demonstrations from such environments are excluded, resulting in a curated dataset of $89500$ high-quality samples used for policy distillation.

\par We distill the teacher policy into the student via IL using behavioral cloning (BC). Demonstration actions are normalized by the action bound $0.3$ so that values lie in the range $[-1,1]$. BC minimizes the discrepancy between teacher actions $a_t$ and student predictions $\pi_s(o_t)$ using the curated dataset. Training is implemented with Robomimic’s BC-Transformer module~\cite{mandlekar2021matters}, which effectively captures temporal dependencies across sequential observations. The student policy uses a $6$-layer transformer with embedding size $512$ and $8$ attention heads, GELU activation and a dropout rate of $0.1$, followed by an MLP head with hidden sizes $[1024,1024]$ to predict actions. We use a context length of $30$, i.e., the input sequence $(o_{t-29},\ldots,o_t)$. The loss is defined as:
\vspace{-0.1cm}
\begin{equation}
L = \| \pi_{s}(o_t) - a_t \|_2^2
\label{eq:imitation_loss}
\end{equation}
\vspace{-0.1cm}
\noindent This objective minimizes the mean squared error between the student prediction $\pi_{s}(o_t)$ and the teacher action $a_t$. Training is conducted on a desktop with an NVIDIA RTX~5080 GPU and completes in approximately $3$~hours. Hyperparameters used for student training are summarized in Table~\ref{tab:transformer_hyper}.

\section{Experiments} \label{sec:Exp}

\begin{table*}[t]
\caption{Ablation study of real-world grasp success rates (\%) on six in-distribution (ID) objects\label{tab:ablation_results}
}
\centering
\small
\resizebox{\textwidth}{!}{%
\begin{tabular}{m{3cm}  
*{5}{>{\centering\arraybackslash}m{2cm}} |
>{\centering\arraybackslash}m{2cm}} 

\toprule
\textbf{Object} & \textbf{Student (Proposed)} & \textbf{Student w/o Stochastic Teacher} & \textbf{Student w/o Curation} & \textbf{Student w/o $r_\text{i}$} & \textbf{RL w/ $o_t$} & \textbf{Object Average}  \\
\midrule
Sphere (M) & \textbf{100.0\%} & 100.0\% & 90.0\% & 0.0\% & 0.0\% & 58.0\% \\
Capsule (M) & \textbf{100.0\%} & 50.0\% & 0.0\% & 0.0\% & 0.0\% & 30.0\% \\
Cuboid A (M) & \textbf{100.0\%} & 100.0\% & 100.0\% & 20.0\% & 90.0\% & 82.0\% \\
Cuboid B (M) & \textbf{100.0\%} & 100.0\% & 100.0\% & 0.0\% & 90.0\% & 78.0\% \\
Cylinder A (M) & \textbf{100.0\%} & 100.0\% & 100.0\% & 100.0\% & 40.0\% & 88.0\% \\
Cylinder B (M) & \textbf{100.0\%} & 100.0\% & 100.0\% & 20.0\% & 0.0\% & 64.0\% \\ 
\midrule
\textbf{Average} & \textbf{100.0\%} & \textbf{91.7\%} & \textbf{81.7\%} & \textbf{23.3\%} & \textbf{36.7\%} & \phantom{0} \\
\bottomrule
\end{tabular}%
}
\vspace{-0.5cm}
\end{table*}

\par This section evaluates the proposed blind grasping policy on real hardware using in-distribution (ID) and out-of-distribution (OOD) objects. We first conduct an ID-based ablation study to analyze the impact of teacher determinism, demonstration curation, and incentive rewards by comparing the proposed student $\pi_s$ with its ablated variants. We then evaluate the generalization capability beyond the simulation-training distribution by benchmarking the proposed student $\pi_s$ on both ID and OOD objects against a partial observation $o_t$ RL policy and a human-teleoperated IL policy, which serve as lower and upper baselines, respectively. The supplementary video illustrates all real-world grasping demonstrations of the proposed system.

\subsection{Ablation Study}

\par We evaluate the effectiveness of our learning strategies through an ablation study on 6 ID objects, which are 3D-printed replicas of the simulation training objects. Quantitative results are summarized in Table~\ref{tab:ablation_results}. The proposed student policy $\pi_{s}$, distilled from a stochastic teacher, is compared against four variants: \textbf{Student w/o Stochastic Teacher} (trained with deterministic teacher demonstrations), \textbf{Student w/o Curation} (trained without demonstration selection criteria), \textbf{Student w/o $r_\text{i}$} (distilled from a teacher trained without the force-incentive reward), and \textbf{RL w/ $o_t$} (trained solely with partial observations). Regarding data collection for the ablated students: the \textbf{Student w/o Stochastic Teacher} uses demonstrations from a deterministic policy, while \textbf{Student w/o Curation} and \textbf{Student w/o $r_\text{i}$} utilize demonstrations generated by executing the teacher for 100 steps per episode without any selection criteria.

\par Experiments are conducted with identical initial conditions, starting from a non-contact state. Each object undergoes 10 trials, where a trial is considered successful if the object is completely lifted and held for at least 1 second within the 10-second duration.

\par The results in Table~\ref{tab:ablation_results} validate the proposed strategy and offer four key implications. First, observation richness is critical. The \textbf{RL w/ $o_t$} policy performs significantly worse than the proposed method, indicating that partial observations alone are insufficient for convergence. In contrast, distilling privileged information ($o_t^{\text{priv}}$) from the teacher allows the student to learn more efficiently, effectively transferring the benefits of rich sensory feedback to the execution phase.

\par Second, proper reward design is essential for distillation. Notably, the \textbf{Student w/o $r_\text{i}$} performs even worse than the \textbf{RL w/ $o_t$} baseline. This suggests that without the incentive reward $r_i$, the teacher fails to generate meaningful force-interaction behaviors to be distilled, proving that a sensor-guided reward is a prerequisite for effective learning under partial observations.

\par Third, performance varies by object geometry. Success rates averaged across variants show a clear hierarchy: Cylinder~A (88.0\%), Cuboid~A (82.0\%), Cuboid~B (78.0\%), Cylinder~B (64.0\%), Sphere (58.0\%), and Capsule (30.0\%). Complete failures occurred for the Capsule (in three variants) and Sphere (in two variants), while Cylinder~A and Cuboid~A had none. This indicates that geometric factors like aspect ratio and surface curvature significantly influence grasping difficulty and must be considered during policy distillation.

\par Finally, combining a stochastic teacher with curation is crucial for robustness. The proposed method consistently outperforms other variants. Specifically, \textbf{Student w/o Stochastic Teacher} and \textbf{Student w/o Curation} both show degradation on difficult objects like the Capsule and Sphere. These results confirm that the stochastic teacher provides trajectory diversity to handle varied shapes, while curation ensures high-quality force interactions, both of which are necessary to learn a generalized and robust student policy.

\subsection{Real-World Grasping Experiments}
\label{sec:real_world}

\begin{figure*}
    \centering
    \includegraphics[width=\textwidth]{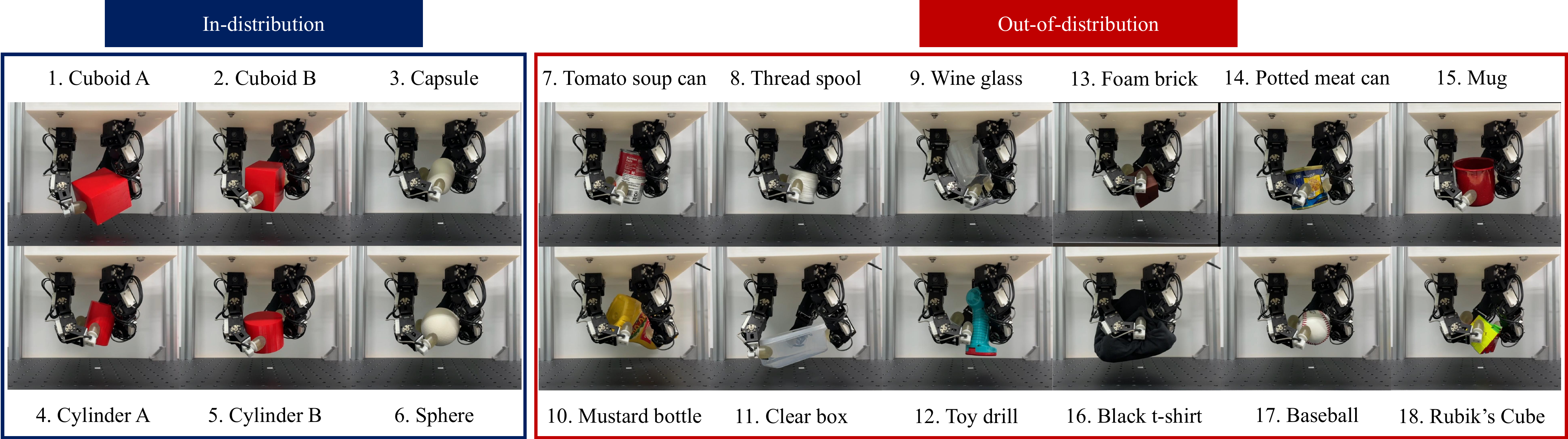}
    \caption{Objects used in real-world evaluation. The evaluation includes 18 objects: six in-distribution (ID) and twelve out-of-distribution (OOD) cases. The ID objects are 3D-printed replicas of the medium-sized training shapes, while the OOD objects are real-world items with diverse shapes, materials, and surface textures that fit within the gripper workspace.}
    \label{real_object}
    \vspace{-0.2cm}
\end{figure*}

\begin{table*}
\caption{Grasping success rate (\%) of the proposed student policy on in-distribution (ID) and out-of-distribution (OOD) objects\label{tab:id_ood_results}}
\centering
\small
\resizebox{\textwidth}{!}{
\begin{tabular}{lcccccc}
\toprule
\textbf{Object (ID)} & \textbf{1. Cuboid A} & \textbf{2. Cuboid B} & \textbf{3. Capsule} & \textbf{4. Cylinder A} & \textbf{5. Cylinder B} & \textbf{6. Sphere} \\
\midrule
Success rate & 100\% & 100\% & 100\% & 100\% & 100\% & 100\% \\
\midrule\midrule
\textbf{Object (OOD)} & \textbf{7. Tomato soup can} & \textbf{8. Thread spool} & \textbf{9. Wine glass} & \textbf{13. Foam brick} & \textbf{14. Potted meat can} & \textbf{15. Mug} \\
\midrule
Success rate & 100\% & 100\% & 100\% & 100\% & 100\% & 100\% \\
\midrule\midrule
\textbf{Object (OOD)} & \textbf{10. Mustard bottle} & \textbf{11. Clear box} & \textbf{12. Toy drill} & \textbf{16. Black T-shirt} & \textbf{17. Baseball} & \textbf{18. Rubik's Cube} \\
\midrule
Success rate & 70\% & 100\% & 100\% & 100\% & 100\% & 100\% \\
\bottomrule
\end{tabular}
} 
\vspace{-0.5cm}
\end{table*}

\par Real-world experiments were conducted using the proposed system with 3D-printed elastic fingertips, which ensure compliance and stable contact. The control loop operates at 20 Hz, mapping uniaxial force and proprioceptive inputs to joint angle changes ($\Delta\theta$). This setup enables zero-shot, vision-free deployment without additional calibration. We evaluated 18 objects (6 ID, 12 OOD) representing diverse shapes, materials, and textures (see Fig.~\ref{real_object}). The ID objects are 3D-printed replicas of training shapes, while OOD objects are real-world items selected to fit within the workspace.

\begin{figure}
    \centering
    \includegraphics[width=8.75cm]     
     {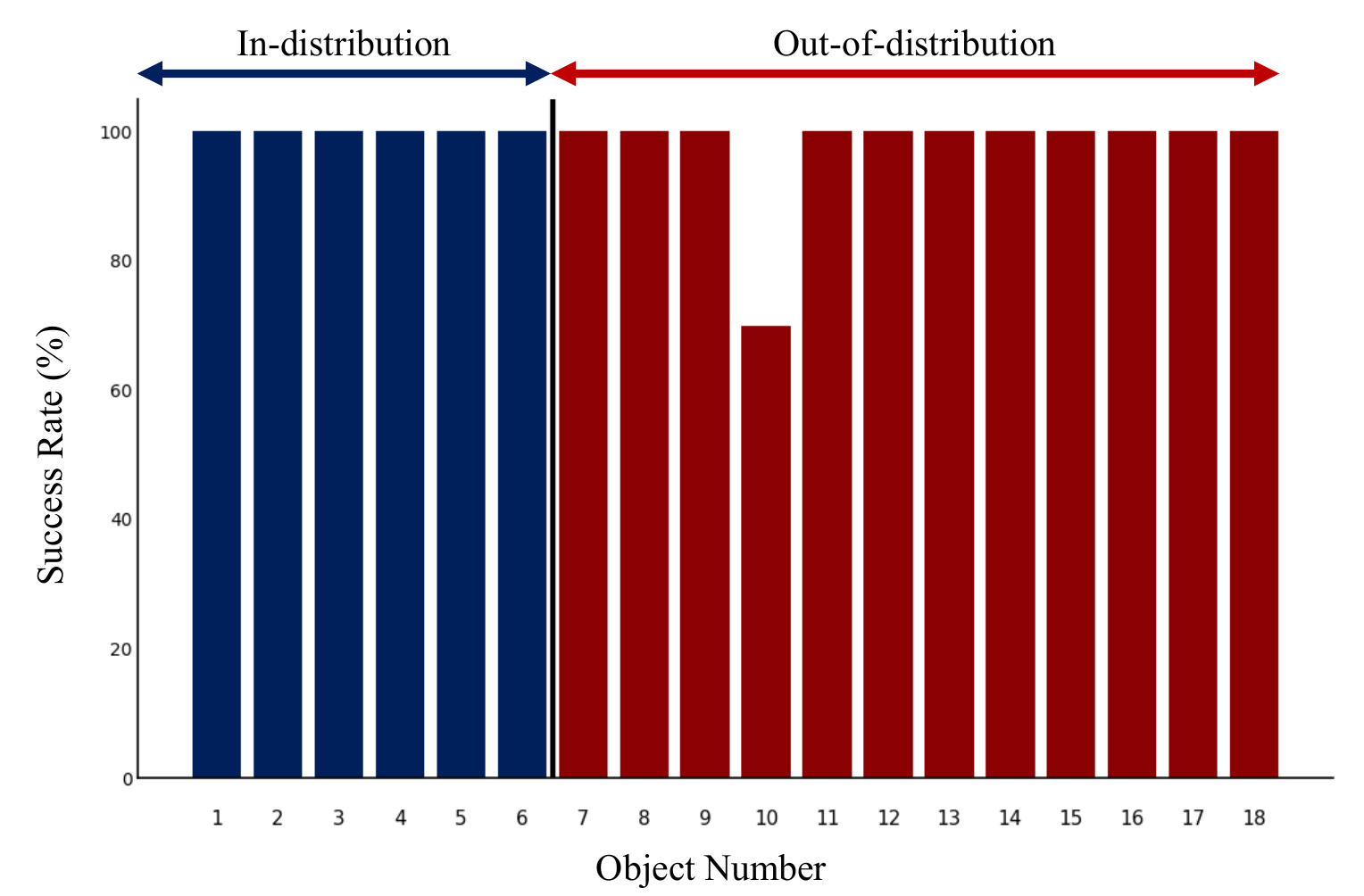}
    \caption{Real-world success rates (\%) of the blind grasping policy: 98.3~$\%$ overall (100.0~$\%$ in-distribution, 97.5~$\%$ out-of-distribution).}
    \label{real_graph}
\vspace{-0.65cm}
\end{figure}

\par As shown in Fig.~\ref{real_graph}, the policy achieved an overall success rate of 98.3\% (100.0\% on ID, 97.5\% on OOD). It demonstrated robust generalization on challenging transparent (Wine glass), soft (Foam brick), and articulated (Rubik’s Cube) objects. The exception was the mustard bottle (70\%), likely due to its length ($21$~cm) exceeding the simulation training range ($10$~cm). We attribute this success to three factors. First, the low-dimensional observation space (proprioception and force) reduces the reality gap, allowing the agent to learn robust patterns from simulation. Second, the policy leverages continuous force feedback to dynamically adapt its grasp to changing contact conditions, thereby enhancing stability and robustness across diverse and challenging objects. Third, the soft elastic fingertips increase friction and contact area, ensuring stable grasp execution.

\subsection{{Comparison Study}}
\label{sec:teleop}

\begin{table}[t]
\caption{Comparison study of real-world grasp success rates (\%) on 18 objects: \textbf{RL w/ $o_t$} vs \textbf{Our Method} vs \textbf{IL Human}\label{tab:teleop_comparison}}

\centering
\small
\begin{tabular}{lccc}
\toprule
\textbf{Object} & \textbf{RL w/ $o_t$} & \textbf{Our Method} & \textbf{IL Human} \\
\midrule
Cuboid A & 90\% & 100\% & 100\% \\
Cuboid B & 90\% & 100\% & 100\% \\
Capsule & 0\% & 100\% & 100\% \\
Cylinder A & 40\% & 100\% & 100\% \\
Cylinder B & 0\% & 100\% & 100\% \\
Sphere & 0\% & 100\% & 100\% \\
\midrule
Tomato soup can & 30\% & 100\% & 100\% \\
Thread spool & 60\% & 100\% & 100\% \\
Wine glass & 50\% & 100\% & 100\% \\
Mustard bottle & 40\% & 70\% & 100\% \\
Clear box & 100\% & 100\% & 100\% \\
Toy drill & 0\% & 100\% & 100\% \\
Foam brick & 0\% & 100\% & 100\% \\
Potted meat can & 60\% & 100\% & 100\% \\
Mug & 50\% & 100\% & 100\% \\
Black T-shirt & 0\% & 100\% & 100\% \\
Baseball & 40\% & 100\% & 100\% \\
Rubik’s Cube & 20\% & 100\% & 100\% \\
\midrule
\textbf{Average} & \textbf{37.2\%} & \textbf{98.3\%} & \textbf{100.0\%}\\
\bottomrule
\end{tabular}
\vspace{-0.5cm}
\end{table}

\begin{figure}
\centering
\includegraphics[width=8.75cm]
{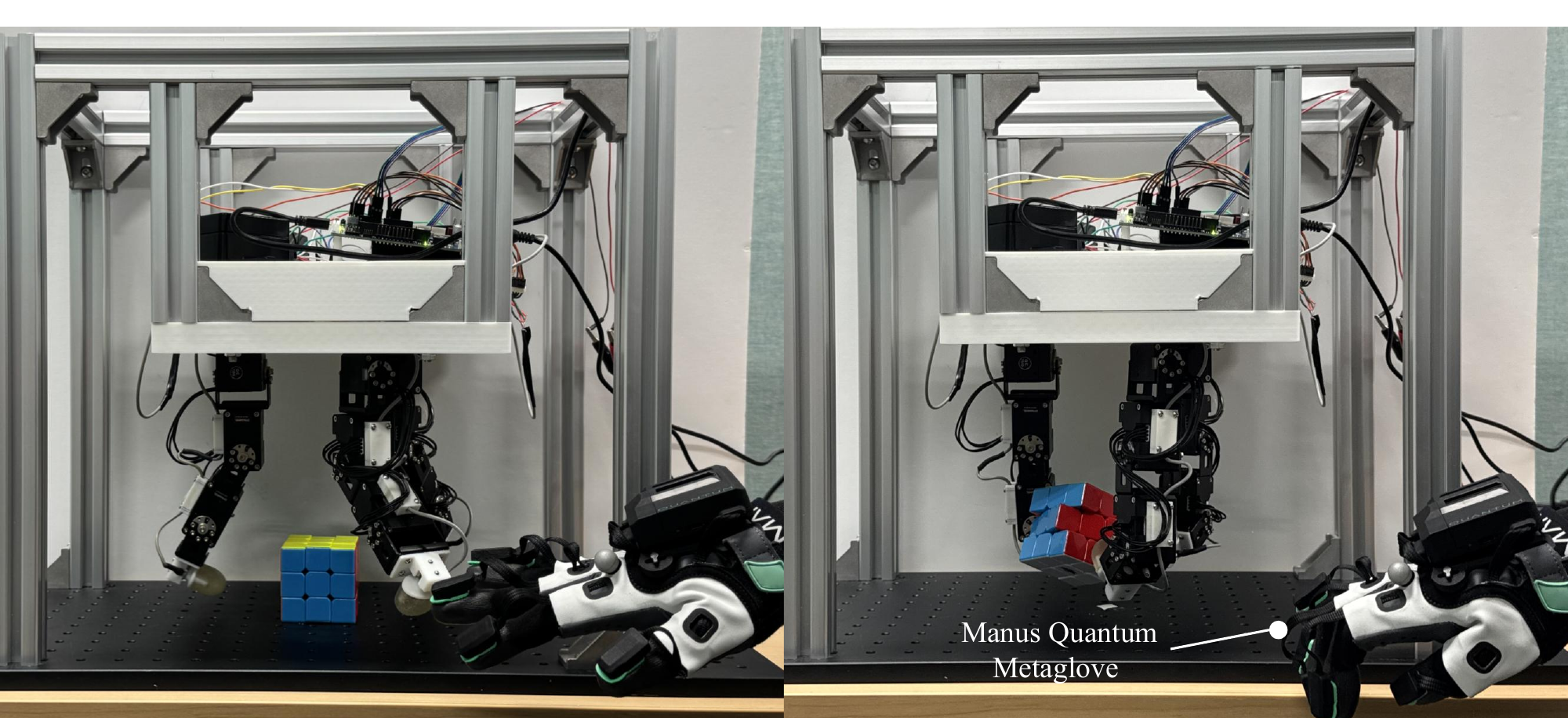}
\caption{Teleoperation setup for collecting human grasp demonstrations.}
\label{fig:demo_exp_setup}
\vspace{-0.65cm}
\end{figure}

\par To benchmark the proposed policy, we conducted a comparative study against two baselines representing performance bounds. As a lower bound, we trained an RL policy with partial observations (\textbf{RL w/ $o_t$}). As an upper bound, we trained an IL policy using high-quality human demonstrations collected via real-world teleoperation (\textbf{IL Human}). \textbf{IL Human} represents an ideal scenario where a human operator utilizes full visual feedback without sim-to-real gaps. All policies were evaluated on the same 18 real-world objects under identical protocols.

\par As shown in Fig.~\ref{fig:demo_exp_setup}, human demonstrations were collected using a Manus Quantum Metaglove~\cite{manus_meta_glove} via a retargeting mapping designed to intuitively align the robot's kinematic structure with human hand movements. The system operated at 20 Hz, recording joint positions and uniaxial force. We define the action as the difference between the glove's target configuration and the robot's current position ($a_t = q_{glove} - q_{robot}$). This formulation enables the low-level PD controller to translate position errors into impedance, implicitly encoding the compliance essential for blind grasping.

\par Data collection followed the real-world validation protocol: fixed initialization and a 10-second duration per episode. To ensure sufficient data quality, we tailored the number of demonstrations based on object difficulty: $60$ for the capsule; $50$ each for the sphere, baseball, thread spool, cuboid A, and clear box; $40$ for the cube; $20$ each for cylinder A and cylinder B; and $10$ for the remaining objects. The context length was set to $30$ to match the student policy.

\par The results highlight the trade-offs between performance and scalability. The lower-bound \textbf{RL w/ $o_t$} achieved only $37.2\%$ success, generalizing poorly across diverse shapes and materials. In contrast, the upper-bound \textbf{IL Human} achieved $100\%$ success on all objects with only 500 demonstrations, confirming the value of high-quality, domain-specific data. However, relying on teleoperation presents practical challenges: extensive force sensor usage introduces drift requiring re-tuning, manual curation is labor-intensive, and human trajectories often contain suboptimal idle segments. Crucially, achieving target performance with Human IL often necessitates an iterative loop of demonstration collection and validation for specific objects, significantly limiting scalability. In contrast, the proposed method offers superior scalability, as expanding to new objects simply requires adding their assets to the simulation environment. Overall, our method achieved $98.3\%$ success—substantially outperforming \textbf{RL w/ $o_t$} and approaching \textbf{IL Human}—while being trained entirely in simulation. This demonstrates that leveraging a teacher–student pipeline with scalable simulation data and objective curation can yield near-human performance without the costs of manual teleoperation.

\section{Discussion} \label{sec:Discussion}

\par Despite robust blind grasping performance, the proposed framework has several limitations. First, without exteroceptive feedback, the policy is sensitive to the initial object position. Large offsets from the gripper center often lead to partial contact, unstable grasps, or objects moving outside the effective workspace, and the current gripper-only policy cannot re-center the object. Extending the system with arm-hand coordination is left for future work. Second, training and evaluation are limited to single-object scenes, making selective target grasping in clutter difficult. We leave multi-object, target-conditioned grasping---potentially guided by a vision-based high-level controller---to future work.

\par In industrial automation, the proposed blind multifinger grasping policy is well suited for general-purpose pick-and-place and bin-picking in logistics and manufacturing. For items within the gripper workspace, it can grasp diverse objects-including irregular and deformable ones-using only onboard proprioception and a uniaxial force signal, without external vision sensors. This vision-free operation is particularly beneficial in factories with changing lighting, transparent/reflective packaging, or frequent occlusions from equipment and nearby robots. The gripper acts as a self-contained module with an embedded policy, enabling straightforward integration with different robot arms via a simple grasp command, similar to a parallel gripper but with a broader grasping capability.

\section{Conclusion} \label{sec:Conclusion}

\par This paper presents a teacher-student framework for blind grasping with a three-fingered gripper. A privileged-observation teacher is trained in simulation via RL and distilled into a transformer student that uses only uniaxial fingertip force sensing and proprioceptive joint states. By avoiding vision and complex tactile sensor arrays, the approach enables a lightweight, low-cost, and robust implementation that is insensitive to lighting and occlusion. Real-world experiments achieve 98.3~$\%$ success over 18 objects (100.0~$\%$ ID, 97.5~$\%$ OOD), demonstrating strong generalization. Ablation and comparison studies show that curated diverse demonstrations and privileged-teacher distillation substantially outperform partial-observation RL and approach a teleoperation-based IL upper bound, supporting deployment in compact automation systems.

 
\bibliographystyle{IEEEtran}
\bibliography{References}

\newpage

 


\begin{IEEEbiography}
    [{\includegraphics[width=1in,height=1.25in,clip,keepaspectratio]{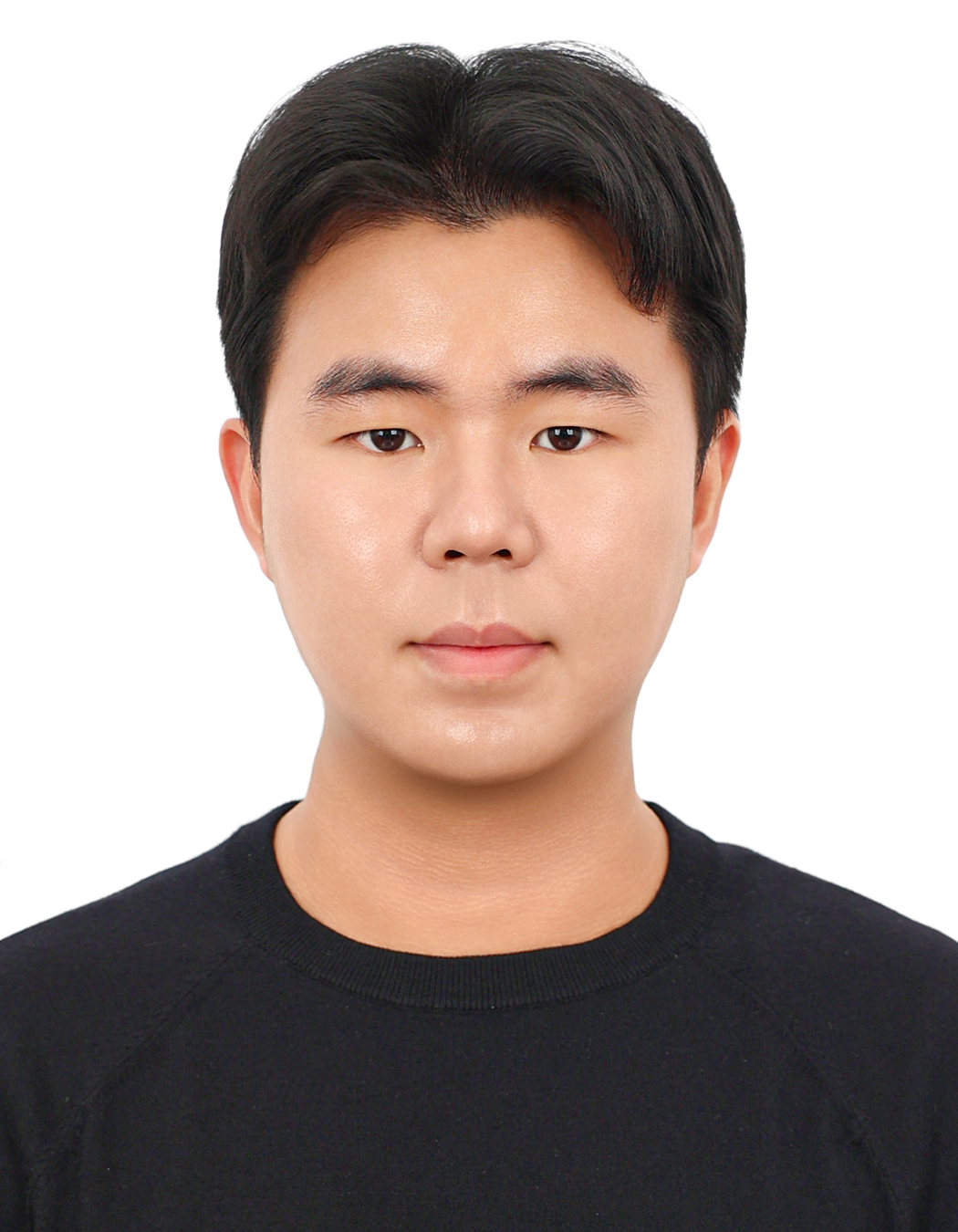}}]{Edgar Lee}
        received a B.S. degree in mechanical engineering from Sogang University, Seoul, South Korea, in 2021. He is currently working toward an integrated Ph.D. degree with the Robotics and Intelligent Mechanisms (RIM) Lab at Sogang University. His research interests are reinforcement learning of robotic systems.
\end{IEEEbiography}

\begin{IEEEbiography}
    [{\includegraphics[width=1in,height=1.25in,clip,keepaspectratio]{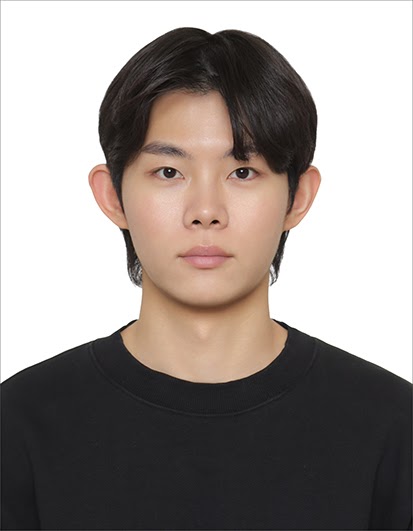}}]{Junho Choi}
      received a B.S. degree in mechanical engineering from Sogang University, Seoul, South Korea, in 2024. He is currently working toward a Master's degree with the Robotics and Intelligent Mechanisms (RIM) Lab at Sogang University. His research interests are reinforcement learning of robotic systems.
\end{IEEEbiography}

\begin{IEEEbiography}
    [{\includegraphics[width=1in,height=1.25in,clip,keepaspectratio]{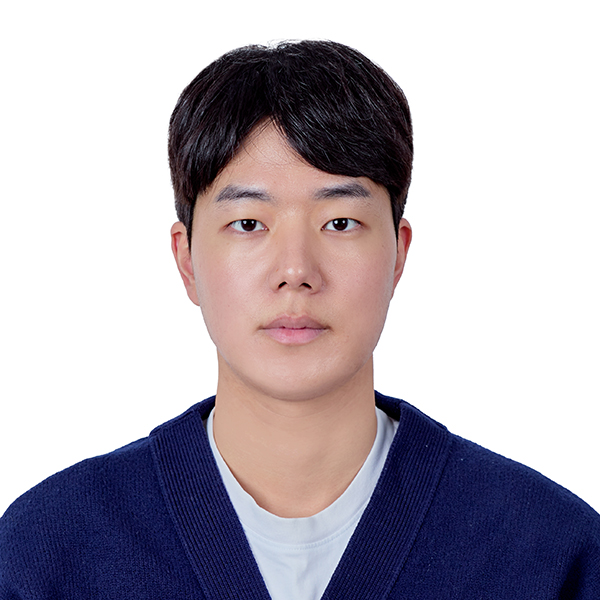}}]{Taemin Kim}
      received a B.S. degree in mechanical engineering from Sogang University, Seoul, South Korea, in 2025. He is currently pursuing a Master's degree at Columbia University. His research interests are reinforcement learning of robotic systems.
\end{IEEEbiography}

\begin{IEEEbiography}
    [{\includegraphics[width=1in,height=1.25in,clip,keepaspectratio]{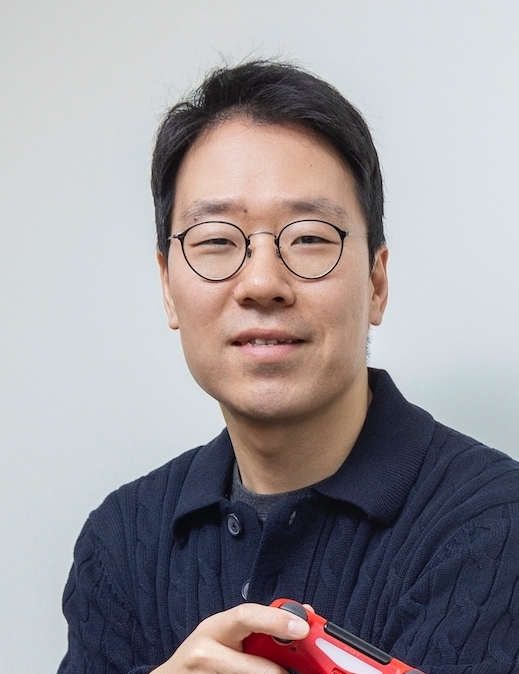}}]{Changjoo Nam}
     received the B.S. and M.S. degrees in electrical engineering from Korea University, Seoul, South Korea, in 2009 and 2011, respectively, and the Ph.D. degree in computer science from Texas A\&M University, College Station, TX, USA, in 2016.
     He is an Associate Professor with the Department of Electronic Engineering at Sogang University since 2022. Before joining Sogang, he was an Assistant Professor at Inha University, Incheon, South Korea and a Senior Research Scientist at the Korea Institute of Science and Technology (KIST), Seoul, South Korea. From 2016 to 2018, he worked with the Robotics Institute, Carnegie Mellon University, Pittsburgh, PA, USA, as a Postdoctoral Fellow. His research interests include task and motion planning and learning-based methods for manipulation and navigation.
\end{IEEEbiography}

\begin{IEEEbiography}
    [{\includegraphics[width=1in,height=1.25in,clip,keepaspectratio]{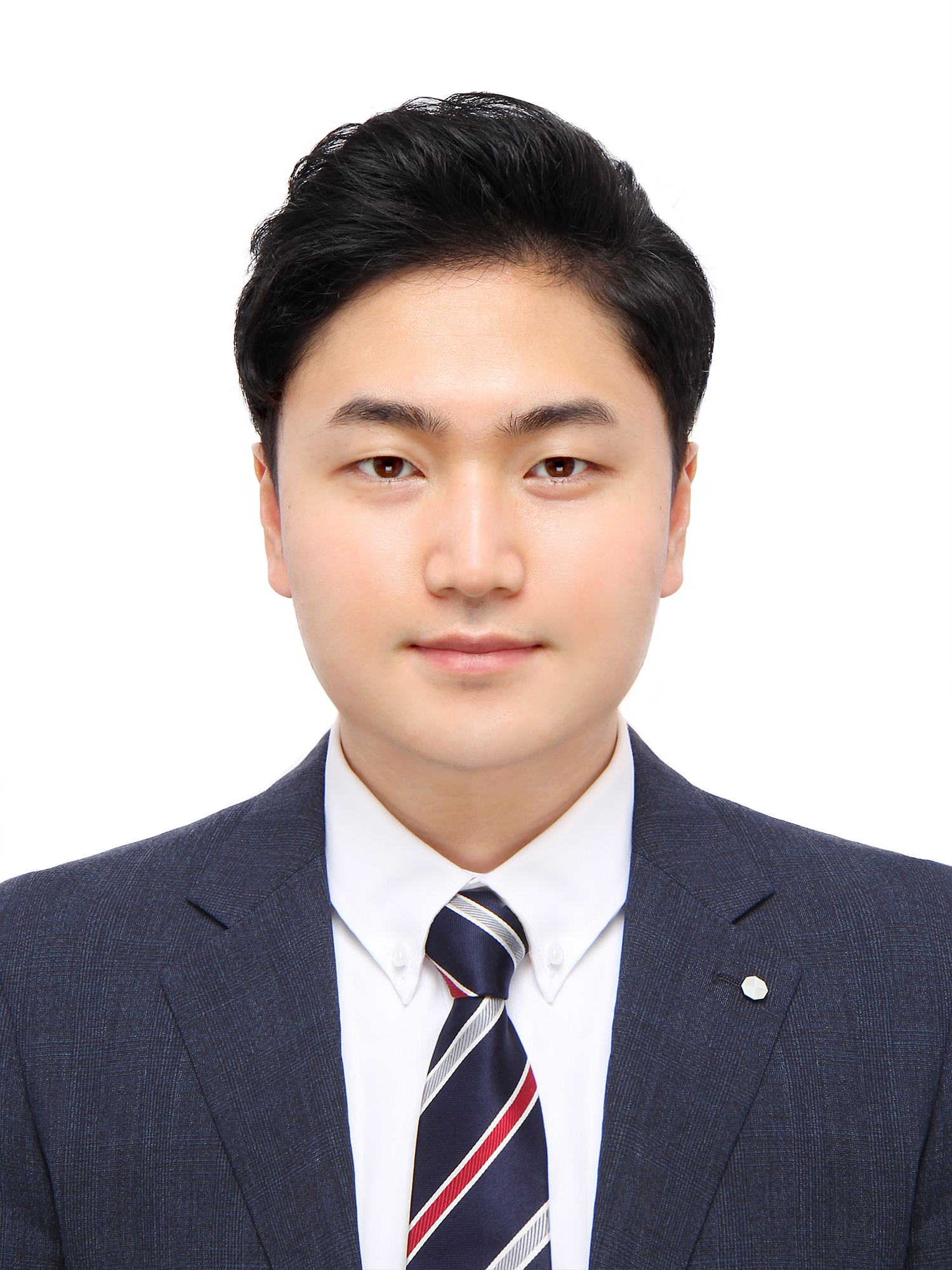}}]{Seokhwan Jeong}
    received a B.S. degree in mechanical engineering from Yonsei University, Seoul, South Korea, in 2013 and M.S. and Ph.D. degrees in mechanical engineering from the Korea Advanced Institute of Science and Technology (KAIST), Daejeon, South Korea in 2015 and 2018, respectively. From 2018 to 2020, he was a postdoctoral researcher in the RoboMed Laboratory in the Department of Biomedical Engineering, Georgia Institute of Technology, Atlanta, GA, USA. He is currently an Associate Professor in the Department of Mechanical Engineering at Sogang University, Seoul, South Korea, and a principal investigator of the Robotics and Intelligent Mechanisms (RIM) Lab (https://rim.sogang.ac.kr/). His research interests include novel actuation mechanisms and designs, robotic hands, prosthetic hand systems, sensor designs, and control.
\end{IEEEbiography}



\vfill

\end{document}